\documentclass{article}
\usepackage{nips13submit_e,times}
\usepackage{url}
\usepackage{amsmath,amsfonts}
\usepackage{graphicx}

\title{Distributed Representations of Words and Phrases \\ and their Compositionality}
\author{
Tomas Mikolov\\
Google Inc.\\
Mountain View \\
\texttt{mikolov@google.com} \\
\And
Ilya Sutskever \\
Google Inc. \\
Mountain View \\
\texttt{ilyasu@google.com} \\
\And
Kai Chen \\
Google Inc. \\
Mountain View \\
\texttt{kai@google.com} \\
\And
Greg Corrado \\
Google Inc. \\
Mountain View \\
\texttt{gcorrado@google.com} \\
\And
Jeffrey Dean \\
Google Inc. \\
Mountain View \\
\texttt{jeff@google.com}}

\nipsfinalcopy

\begin{document}
\maketitle
\begin{abstract}

The recently introduced continuous Skip-gram model is an
efficient method for learning high-quality distributed vector representations that
capture a large number of precise syntactic and semantic word
relationships. In this paper we present several extensions that improve both
the quality of the vectors and the training speed.
By subsampling of the frequent words we obtain significant speedup
and also learn more regular word representations. We also describe a simple
alternative to the hierarchical softmax called negative sampling.

An inherent limitation of word representations is their indifference
to word order and their inability to represent idiomatic phrases.  For
example, the meanings of ``Canada'' and ``Air'' cannot be easily
combined to obtain ``Air Canada''.  Motivated by
this example, we present a simple method for finding
phrases in text, and show that learning good vector
representations for millions of phrases is possible.

\end{abstract}

\section{Introduction}

Distributed representations of words in a vector space
help learning algorithms to achieve
better performance in natural language processing tasks by grouping
 similar words. One of the earliest use of word representations
dates back to 1986 due to Rumelhart, Hinton, and Williams~\cite{nature}.
This idea has since been applied to statistical language modeling with considerable
success~\cite{bengio_lm}. The follow up work includes
applications to automatic speech recognition and machine translation~\cite{Schwenk,mikolov4},
and a wide range of NLP tasks~\cite{collobert,wsabie,socher,bengio_sentiment,turney,turney2,linreg}.

\begin{figure}[t]
\vspace{-0.8cm}
\centering
\centerline{\includegraphics[width=.5\textwidth]{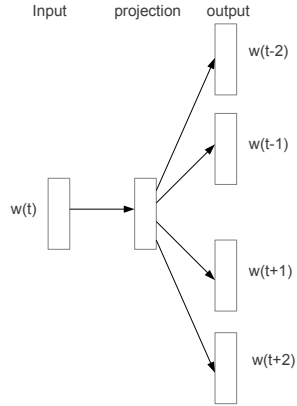}}
\vspace{-3.2cm}
\caption{\small\label{fig:skip}The Skip-gram model architecture. The training objective
  is to learn word vector representations that are good at predicting the nearby words.}
\end{figure}

Recently, Mikolov et al.~\cite{mikolov} introduced the Skip-gram
model, an efficient method for learning high-quality vector
representations of words from large amounts of unstructured text data.
Unlike most of the previously used neural network architectures
for learning word vectors, training of the Skip-gram model (see Figure~\ref{fig:skip})
does not involve dense matrix multiplications. This makes the training
extremely efficient: an optimized single-machine implementation can train
on more than 100 billion words in one day.

The word representations computed using neural networks are
very interesting because the learned vectors explicitly
encode many linguistic regularities and patterns.
Somewhat surprisingly, many of these patterns can be represented
as linear translations.
For example, the result of a vector calculation
vec(``Madrid'') - vec(``Spain'') + vec(``France'') is closer to
vec(``Paris'') than to any other word vector~\cite{linreg,mikolov}.

In this paper we present several extensions of the
original Skip-gram model. We show that subsampling of frequent
words during training results in a significant speedup (around 2x - 10x), and improves
accuracy of the representations of less frequent words.
In addition, we present a simplified variant of Noise Contrastive
Estimation (NCE)~\cite{nce} for training the Skip-gram model that
results in faster training and better vector representations for
frequent words, compared to more complex hierarchical softmax that
was used in the prior work~\cite{mikolov}.

Word representations are limited by their inability to
represent idiomatic phrases that are not compositions of the individual
words. For example, ``Boston Globe'' is a newspaper, and so it is not a
natural combination of the meanings of ``Boston'' and ``Globe''.
Therefore, using vectors to represent
the whole phrases makes the Skip-gram model considerably more
expressive. Other techniques that aim to represent meaning of sentences
by composing the word vectors, such as the
recursive autoencoders~\cite{socher}, would also benefit from using
phrase vectors instead of the word vectors.

The extension from word based to phrase based models is relatively simple.
First we identify a large number of
phrases using a data-driven approach, and then we treat the phrases as
individual tokens during the training. To evaluate the quality of the
phrase vectors, we developed a test set of analogical reasoning tasks that
contains both words and phrases. A typical analogy pair from our test set
is ``Montreal'':``Montreal Canadiens''::``Toronto'':``Toronto Maple Leafs''.
It is considered to have been answered correctly if the
nearest representation to vec(``Montreal Canadiens'') - vec(``Montreal'')
+ vec(``Toronto'') is vec(``Toronto Maple Leafs'').

Finally, we describe another interesting property of the Skip-gram
model. We found that simple vector addition can often produce meaningful
results. For example, vec(``Russia'') + vec(``river'')
is close to vec(``Volga River''), and
vec(``Germany'') + vec(``capital'') is close to vec(``Berlin'').
This compositionality suggests that a non-obvious degree of
language understanding can be obtained by using basic mathematical 
operations on the word vector representations. 

\section{The Skip-gram Model}

The training objective of the Skip-gram model is to find word
representations that are useful for predicting the surrounding words in a sentence
or a document.
More formally, given a sequence of training words $w_1, w_2,
w_3,\ldots,w_T$, the objective of the Skip-gram model is to maximize
the average log probability
\begin{equation}
\frac1T\sum_{t=1}^T\sum_{-c\leq j\leq c, j\neq 0} \log p(w_{t+j}|w_t)
\end{equation}
where $c$ is the size of the training context (which can be a function
of the center word $w_t$). Larger $c$ results in more
training examples and thus can lead to a higher accuracy, at the
expense of the training time. The basic Skip-gram formulation defines
$p(w_{t+j}|w_t)$ using the softmax function:
\begin{equation}
p(w_O|w_I) = \frac{\exp\left({v'_{w_O}}^\top v_{w_I}\right)}
{\sum_{w=1}^{W}\exp\left({v'_w}^\top v_{w_I}\right)}
\end{equation}
where $v_w$ and $v'_w$ are the ``input'' and ``output'' vector representations
of $w$, and $W$ is the number of words in the vocabulary. This
formulation is impractical because the cost of computing $\nabla \log
p(w_O|w_I)$ is proportional to $W$, which is often large
($10^5$--$10^7$ terms).

\subsection{Hierarchical Softmax}
\label{sec:hsoft}

A computationally efficient approximation of the full softmax is the hierarchical softmax.
In the context of neural network language models, it was first
introduced by Morin and Bengio~\cite{hsoft_first}. The main
advantage is that instead of evaluating $W$ output nodes in the neural network to obtain
the probability distribution, it is needed to evaluate only about $\log_2(W)$ nodes.

The hierarchical softmax uses a binary tree representation of the output layer
with the $W$ words as its leaves and, for each
node, explicitly represents the relative probabilities of its child
nodes. These define a random walk that assigns probabilities to words.

More precisely, each word $w$ can be reached by an appropriate path
from the root of the tree. Let $n(w,j)$ be the $j$-th node on the
path from the root to $w$, and let $L(w)$ be the length of this path,
so $n(w,1)=\mathrm{root}$ and $n(w,L(w))=w$.  In addition, for any
inner node $n$, let $\mathrm{ch}(n)$ be an arbitrary fixed child of
$n$ and let $[\![x]\!]$ be 1 if $x$ is true and -1 otherwise.
Then the hierarchical softmax defines $p(w_O|w_I)$ as follows:
\begin{equation}
p(w|w_I) = \prod_{j=1}^{L(w)-1}
\sigma\left([\![n(w,j+1)=\mathrm{ch}(n(w,j))]\!]\cdot {v'_{n(w,j)}}^\top
v_{w_I}\right)
\end{equation}
where $\sigma(x)=1/(1+\exp(-x))$. It can be verified that
$\sum_{w=1}^W p(w|w_I)=1$. This implies that
the cost of computing $\log p(w_O|w_I)$ and $\nabla
\log p(w_O|w_I)$ is proportional to $L(w_O)$, which on average is no greater
than $\log W$. Also, unlike the standard softmax formulation of the Skip-gram
which assigns two representations $v_w$ and $v'_w$ to each word $w$, the
hierarchical softmax formulation has
one representation $v_w$ for each word $w$ and one representation $v'_n$
for every inner node $n$ of the binary tree.

The structure of the tree used by the hierarchical softmax has
a considerable effect on the performance. Mnih and Hinton
explored a number of methods for constructing the tree structure
and the effect on both the training time and the resulting model accuracy~\cite{mnih}.
In our work we use a binary Huffman tree, as it assigns short codes to the frequent words
which results in fast training. It has been observed before that grouping words together
by their frequency works well as a very simple speedup technique for the neural
network based language models~\cite{mikolov3,mikolov}.

\begin{figure}[t]
\centering
\centerline{\includegraphics[width=0.9\textwidth]{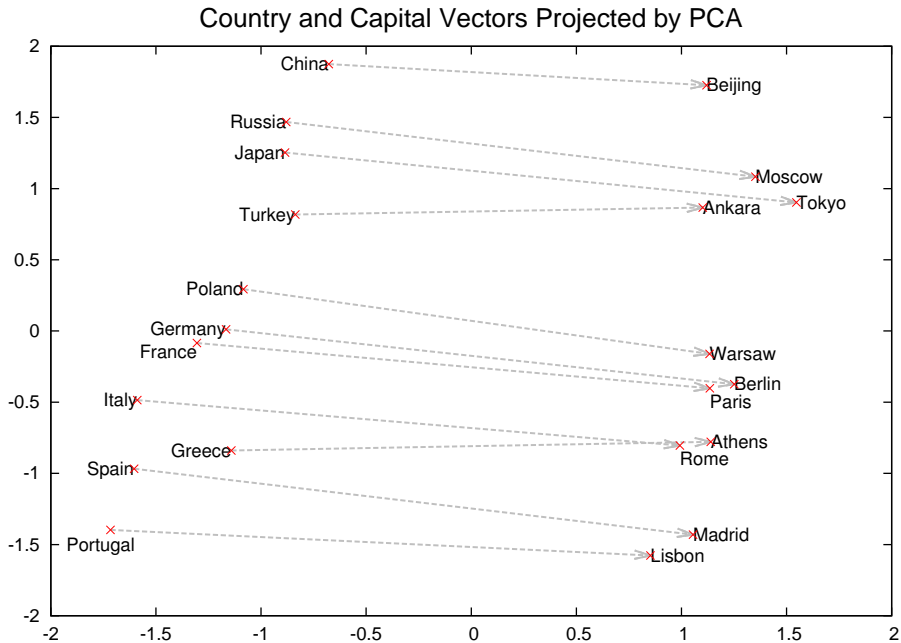}}
\caption{\small Two-dimensional PCA projection of the
  1000-dimensional Skip-gram vectors of countries and their capital
  cities. The figure illustrates ability of the model to automatically organize
  concepts and learn implicitly the relationships between them, as during the
  training we did not provide any supervised information about what a capital
  city means.}
\end{figure}

\subsection{Negative Sampling}

An alternative to the hierarchical softmax is Noise Contrastive
Estimation (NCE), which was introduced by Gutmann and Hyvarinen~\cite{nce}
and applied to language modeling by Mnih and Teh~\cite{mnih-nce}.
NCE posits that a good model should be able to
differentiate data from noise by means of logistic regression. This is
similar to hinge loss used by Collobert and Weston~\cite{collobert} who trained
the models by ranking the data above noise.

While NCE can be shown to approximately maximize the log
probability of the softmax, the Skip-gram model is only concerned with learning
high-quality vector representations, so we are free to simplify NCE as
long as the vector representations retain their quality. We define Negative sampling (NEG)
by the objective
 \begin{equation}
   \log \sigma({v'_{w_O}}^\top v_{w_I}) + \sum_{i=1}^k\mathbb E_{w_i\sim
    P_n(w)}\left[\log \sigma(-{v'_{w_i}}^\top v_{w_I})\right]
 \end{equation}
which is used to replace every $\log P(w_O|w_I)$ term in the Skip-gram objective.
Thus the task is to distinguish the target word
$w_O$ from draws from the noise distribution $P_n(w)$ using logistic regression,
where there are $k$ negative
samples for each data sample. Our experiments indicate that values of $k$
in the range 5--20 are useful for small training datasets, while for large datasets
the $k$ can be as small as 2--5. The main difference between the Negative sampling and NCE is that NCE
needs both samples and the numerical probabilities of the noise distribution,
while Negative sampling uses only samples. And while NCE approximately maximizes the log probability
of the softmax, this property is not important for our application.

Both NCE and NEG have the noise distribution $P_n(w)$ as
a free parameter. We investigated a number of choices for $P_n(w)$
and found that the unigram distribution $U(w)$ raised to the $3/4$rd
power (i.e., $U(w)^{3/4}/Z$) outperformed significantly the unigram
and the uniform distributions, for both NCE and NEG on every task we tried
including language modeling (not reported here).

\subsection{Subsampling of Frequent Words}

In very large corpora, the most frequent words can easily occur hundreds of millions
of times (e.g., ``in'', ``the'', and ``a''). Such words usually
provide less information value than the rare words. For example, while the
Skip-gram model benefits from observing the co-occurrences of ``France'' and
``Paris'', it benefits much less from observing the frequent co-occurrences of ``France''
and ``the'', as nearly every word co-occurs frequently within a sentence
with ``the''. This idea can also be applied in the opposite
direction; the vector representations of frequent words do not change
significantly after training on several million examples.

To counter the imbalance between the rare and frequent words, we used a
simple subsampling approach: each word $w_i$ in the training set is
discarded with probability computed by the formula
\begin{equation}
P(w_i) = 1-\sqrt{\frac{t}{f(w_i)}}
\end{equation}
where $f(w_i)$ is the frequency of word $w_i$ and $t$ is a chosen
threshold, typically around $10^{-5}$.  We chose this subsampling
formula because it aggressively subsamples words whose frequency is
greater than $t$ while preserving the ranking of the frequencies.
Although this subsampling formula was chosen heuristically, we found
it to work well in practice. It accelerates learning and even significantly improves
the accuracy of the learned vectors of the rare words, as will be shown in the following sections.

\section{Empirical Results}

In this section we evaluate the Hierarchical Softmax (HS), Noise Contrastive Estimation,
Negative Sampling, and subsampling of the training words. We used
the analogical reasoning task\footnote{\url{code.google.com/p/word2vec/source/browse/trunk/questions-words.txt}}
introduced by Mikolov et al.~\cite{mikolov}.
The task consists of analogies such as ``Germany'' : ``Berlin'' :: ``France'' : ?,
which are solved by finding a vector $\mathbf{x}$
such that vec($\mathbf{x}$) is closest to
vec(``Berlin'') - vec(``Germany'') + vec(``France'') according to the
cosine distance (we discard the input words from the search).
This specific example is considered to have been
answered correctly if $\mathbf{x}$ is ``Paris''.  The task has
two broad categories: the syntactic analogies (such as
``quick'' : ``quickly'' :: ``slow'' : ``slowly'') and the semantic analogies, such
as the country to capital city relationship.

For training the Skip-gram models, we have used a large dataset
consisting of various news articles (an internal Google dataset with one billion words).
We discarded from the vocabulary all words that occurred
less than 5 times in the training data, which resulted in a vocabulary of size 692K.
The performance of various Skip-gram models on the word
analogy test set is reported in Table~\ref{table:performance_1}.
The table shows that Negative Sampling
outperforms the Hierarchical Softmax on the analogical
reasoning task, and has even slightly better performance than the Noise Contrastive Estimation.
The subsampling of the frequent words improves the training speed several times
and makes the word representations significantly more accurate.


It can be argued that the linearity of the skip-gram model makes its vectors
more suitable for such linear analogical reasoning, but the results of
Mikolov et al.~\cite{mikolov} also show that the vectors learned by the
standard sigmoidal recurrent neural networks (which are highly non-linear)
improve on this task significantly as the amount of the training data increases,
suggesting that non-linear models also have a preference for a linear
structure of the word representations.

\begin{table}
\small
  \begin{center}
  \begin{tabular}{|c|c|c c |c|}
    \hline 
    Method & Time [min]  & Syntactic [\%] &  Semantic [\%] & Total accuracy [\%] \\
    \hline
    NEG-5  & 38          & 63        &  54       & 59    \\
    NEG-15 & 97          & 63        &  58       & {\bf 61}    \\
    HS-Huffman & 41      & 53        &  40       & 47    \\
    NCE-5  & 38          & 60        &  45       & 53    \\
    \hline
    \multicolumn{5}{|c|}{The following results use $10^{-5}$ subsampling}   \\
    \hline
    NEG-5  & 14          & 61        & 58        & 60    \\
    NEG-15 & 36          & 61        & 61        & {\bf 61}    \\
    HS-Huffman & 21      & 52        & 59        & 55     \\
    \hline
  \end{tabular}
  \end{center}
  \caption{\label{table:performance_1} Accuracy of
    various Skip-gram 300-dimensional models on the
    analogical reasoning task as defined in~\cite{mikolov}.
    NEG-$k$ stands for Negative Sampling with $k$
    negative samples for each positive sample; NCE stands for Noise Contrastive Estimation and
    HS-Huffman stands for the Hierarchical Softmax with the frequency-based Huffman
    codes.}
\end{table}

\section{Learning Phrases}

As discussed earlier, many phrases have a
meaning that is not a simple composition of the meanings of its individual
words. To learn vector representation for phrases, we first
find words that appear frequently together, and infrequently
in other contexts. For example, ``New York Times'' and
``Toronto Maple Leafs'' are replaced by unique tokens in the training data,
while a bigram ``this is'' will remain unchanged.

This way, we can form many reasonable phrases without greatly increasing the size
of the vocabulary; in theory, we can train the Skip-gram model
using all n-grams, but that would
be too memory intensive. Many techniques have been previously developed
to identify phrases in the text;
however, it is out of scope of our work to compare them. We decided to use
a simple data-driven approach, where phrases are formed
based on the unigram and bigram counts, using
\begin{equation}
\textrm{score}(w_i, w_j) = \frac{\textrm{count}(w_iw_j) - \delta}{\textrm{count}(w_i)\times \textrm{count}(w_j)}.
\end{equation}
The $\delta$ is used as a discounting coefficient and prevents too many
phrases consisting of very infrequent words to be formed.
The bigrams with score above the chosen threshold are then used as phrases.
Typically, we run 2-4 passes over the training data with decreasing
threshold value, allowing longer phrases that consists of several words to be formed.
We evaluate the quality of the phrase representations using a new analogical
reasoning task that involves phrases. Table~\ref{table:phrase_example} shows
examples of the five categories of analogies used in this task. This dataset is publicly available
on the web\footnote{\url{code.google.com/p/word2vec/source/browse/trunk/questions-phrases.txt}}.

\begin{table}
\small
\begin{center}
\begin{tabular}{|c c||c c|}
\hline
\multicolumn{4}{|c|}{Newspapers}\\
\hline
New York & New York Times & Baltimore & Baltimore Sun \\
San Jose & San Jose Mercury News & Cincinnati & Cincinnati Enquirer \\
\hline
\multicolumn{4}{|c|}{NHL Teams}\\
\hline
Boston & Boston Bruins & Montreal & Montreal Canadiens \\
Phoenix & Phoenix Coyotes & Nashville & Nashville Predators \\
\hline
\multicolumn{4}{|c|}{NBA Teams}\\
\hline
Detroit & Detroit Pistons & Toronto & Toronto Raptors \\
Oakland & Golden State Warriors & Memphis & Memphis Grizzlies \\
\hline
\multicolumn{4}{|c|}{Airlines}\\
\hline
Austria & Austrian Airlines & Spain & Spainair \\
Belgium & Brussels Airlines & Greece & Aegean Airlines \\
\hline
\multicolumn{4}{|c|}{Company executives}\\
\hline
Steve Ballmer & Microsoft & Larry Page & Google \\
Samuel J. Palmisano & IBM & Werner Vogels & Amazon \\
\hline
\end{tabular}
\end{center}
\caption{
\label{table:phrase_example} Examples of the
analogical reasoning task for phrases (the full test set has 3218 examples).
The goal is to compute the fourth phrase using the first three. Our best model achieved an accuracy of 72\% on this dataset.}
\end{table}

\subsection{Phrase Skip-Gram Results}
\label{sec:phrase_results}

Starting with the same news data as in the previous experiments,
we first constructed the phrase based training corpus and then we trained several
Skip-gram models using different hyper-parameters. As before, we used vector
dimensionality 300 and context size 5. This
setting already achieves good performance on the phrase
dataset, and allowed us to quickly compare the Negative Sampling
and the Hierarchical Softmax, both with and without subsampling
of the frequent tokens. The results are summarized in Table~\ref{table:phrase_basic_results}.
\begin{table}[b]
\small
\begin{center}
  \begin{tabular}{|c|c|c|c|}
    \hline
    Method & Dimensionality & No subsampling [\%] & $10^{-5}$ subsampling [\%] \\
    \hline
    NEG-5  &    300         & 24                  & 27                      \\
    NEG-15 &    300         & 27                  & 42                      \\
    HS-Huffman & 300        & 19                  & {\bf 47}                \\
    \hline
  \end{tabular}
\end{center}
\caption{
\label{table:phrase_basic_results}
Accuracies of the Skip-gram models on the phrase analogy dataset. The models were
trained on approximately one billion words from the news dataset.}
\end{table}

The results show that while Negative Sampling achieves a respectable
accuracy even with $k=5$, using $k=15$ achieves considerably better
performance. Surprisingly, while we found the Hierarchical Softmax to
achieve lower performance when trained without subsampling,
it became the best performing method when we
downsampled the frequent words. This shows that the subsampling
can result in faster training and can also improve accuracy, at least in some cases.

To maximize the accuracy on the phrase analogy task, we increased
the amount of the training data by using a dataset with about 33 billion words. We
used the hierarchical softmax, dimensionality of 1000, and
the entire sentence for the context.
This resulted in a model that reached an accuracy of {\bf 72\%}. We achieved lower accuracy
66\% when we reduced the size of the training dataset to 6B words, which suggests
that the large amount of the training data is crucial.

To gain further insight into how different the representations learned by different
models are, we did inspect manually the nearest neighbours of infrequent phrases
using various models. In Table~\ref{table:comparison}, we show a sample of such comparison.
Consistently with the previous results, it seems that the best representations of
phrases are learned by a model with the hierarchical softmax and subsampling.
\begin{table}
\small
\begin{center}
\begin{tabular}{|c|c|c|}
\hline
                  & NEG-15 with $10^{-5}$ subsampling  &         HS with $10^{-5}$ subsampling \\
\hline
 Vasco de Gama    &   Lingsugur              &            Italian explorer \\
Lake Baikal       &   Great Rift Valley      &            Aral Sea \\
 Alan Bean        &   Rebbeca Naomi          &            moonwalker \\
Ionian Sea        &   Ruegen                 &            Ionian Islands \\
chess master      &   chess grandmaster      &            Garry Kasparov   \\
\hline
\end{tabular}
\end{center}
\caption{\label{table:comparison} Examples of the closest entities to the given short phrases, using two different models.}
\end{table}

\section{Additive Compositionality}

We demonstrated that the word and phrase representations learned by the Skip-gram
model exhibit a linear structure that makes it possible to perform
precise analogical reasoning using simple vector arithmetics.
Interestingly, we found that the Skip-gram representations exhibit
another kind of linear structure that makes it possible to meaningfully combine
words by an element-wise addition of their vector representations.
This phenomenon is illustrated in Table~\ref{table:composition}.

\begin{table}
\small
  \begin{center}
  \begin{tabular}{|c|c|c|c|c|}
    \hline
    Czech + currency  & Vietnam + capital & German + airlines      &       Russian + river   &  French + actress             \\
    \hline
    koruna            & Hanoi             & airline Lufthansa      &       Moscow            &  Juliette Binoche     \\
    Check crown       & Ho Chi Minh City  & carrier Lufthansa      &       Volga River       &  Vanessa Paradis      \\
    Polish zolty      & Viet Nam          & flag carrier Lufthansa &       upriver           &  Charlotte Gainsbourg \\
    CTK               & Vietnamese        & Lufthansa              &       Russia            &  Cecile De                    \\
    \hline
  \end{tabular}
  \end{center}
\caption{\label{table:composition} Vector compositionality using element-wise addition. Four closest tokens to the sum
of two vectors are shown, using the best Skip-gram model.}
\end{table}

The additive property of the vectors can be explained by inspecting the
training objective.
The word vectors are in a linear relationship with the inputs
to the softmax nonlinearity.
As the word vectors are trained
to predict the surrounding words in the sentence, the vectors
can be seen as representing the distribution of the context in which a word
appears.  These values are related logarithmically to the probabilities
computed by the output layer, so the sum of two word vectors is related to
the product of the two context distributions. 
The product works here as the AND function: words that are
assigned high probabilities by both word vectors will have high probability, and
the other words will have low probability.
Thus, if ``Volga River'' appears frequently in the same sentence together
with the words ``Russian'' and ``river'', the sum of these two word vectors
will result in such a feature vector that is close to the vector of ``Volga River''.

\section{Comparison to Published Word Representations}

Many authors who previously worked on the neural network based representations of words have published their resulting
models for further use and comparison: amongst the most well known authors
are Collobert and Weston~\cite{collobert}, Turian et al.~\cite{turian},
and Mnih and Hinton~\cite{mnih}. We downloaded their word vectors from
the web\footnote{\url{http://metaoptimize.com/projects/wordreprs/}}. 
Mikolov et al.~\cite{mikolov} have already evaluated these word representations on the word analogy task,
where the Skip-gram models achieved the best performance with a huge margin.

To give more insight into the difference of the quality of the learned
vectors, we provide empirical comparison by showing the nearest neighbours of infrequent
words in Table~\ref{table:closest_words}. These examples show that the big Skip-gram model trained on a large
corpus visibly outperforms all the other models in the quality of the learned representations.
This can be attributed in part to the fact that this model
has been trained on about 30 billion words, which is about two to three orders of magnitude more data than
the typical size used in the prior work. Interestingly, although the training set is much larger,
the training time of the Skip-gram model is just a fraction
of the time complexity required by the previous model architectures.

\begin{table}
\small
  \begin{center}
  \begin{tabular}{|c||c|c|c|c|c|}
    \hline
    Model                 & Redmond & Havel       & ninjutsu & graffiti   & capitulate \\
    (training time)       &         &             &          &            &            \\
    \hline
    \hline
    Collobert (50d)       & conyers & plauen      & reiki    & cheesecake & abdicate   \\
    (2 months)            & lubbock & dzerzhinsky & kohona   & gossip     & accede     \\
                          & keene   & osterreich  & karate   & dioramas   & rearm      \\
    \hline
    Turian (200d)         & McCarthy & Jewell     & -        & gunfire    & -      \\
    (few weeks)           & Alston   & Arzu       & -        & emotion    & -      \\
                          & Cousins  & Ovitz      & -        & impunity   & -      \\
    \hline
    Mnih (100d)           & Podhurst & Pontiff    & -        & anaesthetics & Mavericks        \\
    (7 days)              & Harlang  & Pinochet   & -        & monkeys      & planning         \\
                          & Agarwal  & Rodionov   & -        & Jews         & hesitated        \\
    \hline
    Skip-Phrase           & Redmond Wash.         & Vaclav Havel            & ninja         & spray paint   & capitulation         \\
    (1000d, 1 day)        & Redmond Washington    & president Vaclav Havel  & martial arts  & grafitti      & capitulated          \\
                          & Microsoft             & Velvet Revolution       & swordsmanship & taggers       & capitulating      \\
    \hline
  \end{tabular}
  \end{center}
\caption{\label{table:closest_words} Examples of the closest tokens given various
well known models and the Skip-gram model trained on phrases using over 30 billion training words. An empty cell
means that the word was not in the vocabulary.}
\normalsize
\end{table}

\section{Conclusion}

This work has several key contributions.  We show how to train distributed
representations of words and phrases with the Skip-gram model and demonstrate that these
representations exhibit linear structure that makes precise analogical reasoning
possible. The techniques introduced in this paper can be used also for training
the continuous bag-of-words model introduced in~\cite{mikolov}.

We successfully trained models on several orders of magnitude more data than
the previously published models, thanks to the computationally efficient model architecture.
This results in a great improvement in the quality of the learned word and phrase representations,
especially for the rare entities.
We also found that the subsampling of the frequent
words results in both faster training and significantly better representations of uncommon
words. Another contribution of our paper is the Negative sampling algorithm,
which is an extremely simple training method
that learns accurate representations especially for frequent words.

The choice of the training algorithm and the hyper-parameter selection
is a task specific decision, as we found that different problems have
different optimal hyperparameter configurations. In our experiments,
the most crucial decisions that affect the performance are the choice of
the model architecture, the size of the vectors, the subsampling rate,
and the size of the training window.

A very interesting result of this work is that the word vectors
can be somewhat meaningfully combined using
just simple vector addition. Another approach for learning representations
of phrases presented in this paper is to simply represent the phrases with a single
token. Combination of these two approaches gives a powerful yet simple way
how to represent longer pieces of text, while having minimal computational
complexity. Our work can thus be seen as complementary to the existing
approach that attempts to represent phrases using recursive
matrix-vector operations~\cite{socher2}.

We made the code for training the word and phrase vectors based on the techniques
described in this paper available as an open-source project\footnote{\url{code.google.com/p/word2vec}}.

\small
\bibliographystyle{plain}
\bibliography{phrase_rep}

\end{document}